\documentclass{article}

\PassOptionsToPackage{numbers, compress}{natbib}
\usepackage[final]{nips_2016}

\usepackage[utf8]{inputenc} 
\usepackage[T1]{fontenc}    
\usepackage{hyperref}       
\usepackage{url}            
\usepackage{booktabs}       
\usepackage{amsfonts}       
\usepackage{nicefrac}       
\usepackage{microtype}      

\usepackage{graphicx}
\usepackage{multirow}
\usepackage{hhline}
\usepackage{stfloats}
\usepackage{setspace}
\usepackage[table]{xcolor}

\definecolor{maroon}{cmyk}{0,0.87,0.68,0.32}

\title{Neural Semantic Encoders}

\author{
  Tsendsuren Munkhdalai \\
  University of Massachusetts, MA, USA\\
  \texttt{tsendsuren.munkhdalai@umassmed.edu} \\
  \And
  Hong Yu \\
  University of Massachusetts, MA, USA \\
   \texttt{hong.yu@umassmed.edu}  \\
}

\begin{document}

\maketitle

\begin{abstract}
We present a memory augmented neural network for natural language understanding: Neural Semantic Encoders. NSE is equipped with a novel memory update rule and has a variable sized encoding memory that evolves over time and maintains the understanding of input sequences  through \textit{read}, \textit{compose} and \textit{write} operations. NSE can also access\footnote{By access we mean changing the memory states by the \textit{read}, \textit{compose} and \textit{write} operations.} multiple and shared memories. 
In this paper, we demonstrated the effectiveness and the flexibility of NSE on five different natural language tasks: natural language inference, question answering, sentence classification, document sentiment analysis and machine translation where NSE achieved state-of-the-art performance when evaluated on publically available benchmarks. For example, our shared-memory model showed an encouraging result on neural machine translation, improving an attention-based baseline by approximately 1.0 BLEU.
\end{abstract}

\section{Introduction}

Recurrent neural networks (RNNs) have been successful for modeling sequences \cite{elman:90}. Particularly, RNNs equipped with internal short memories, such as long short-term memories (LSTM) \cite{hochreiter:97} have achieved a notable success in sequential tasks \cite{cho2014learning,vinyals:15a}. LSTM is powerful because it learns to control its short term memories. However, the short term memories in LSTM are a part of the training parameters. This imposes some practical difficulties in training and modeling long sequences with LSTM.

Recently several studies have explored ways of extending the neural networks with an external memory \cite{graves2014neural,weston:15,grefenstette2015learning}. Unlike LSTM, the short term memories and the training parameters of such a neural network are no longer coupled and can be adapted. In this paper we propose a novel class of memory augmented neural networks called Neural Semantic Encoders (NSE) for natural language understanding. NSE offers several desirable properties. NSE has a variable sized encoding memory which allows the model to access entire input sequence during the reading process; therefore efficiently delivering long-term dependencies over time. The encoding memory evolves over time and maintains the memory of the input sequence through \textit{read}, \textit{compose} and \textit{write} operations. NSE sequentially processes the input and supports word compositionality inheriting both temporal and hierarchical nature of human language. NSE can read from and write to a set of relevant encoding memories simultaneously or multiple NSEs can access a shared encoding memory effectively supporting knowledge and representation sharing. NSE is flexible, robust and suitable for practical NLU tasks and can be trained easily by any gradient descent optimizer.

We evaluate NSE on five different real tasks. For four of them, our models set new state-of-the-art results. Our results suggest that a NN model with the shared memory between encoder and decoder is a promising approach for sequence transduction problems such as machine translation and abstractive summarization. In particular, we observe that the attention-based neural machine translation can be further improved by shared-memory models. We also analyze memory access pattern and compositionality in NSE and show that our model captures semantic and syntactic structures of input sentence. 

\section{Related Work}

\begin{figure*}[tp]
    \centering
        \includegraphics[width=1\textwidth]{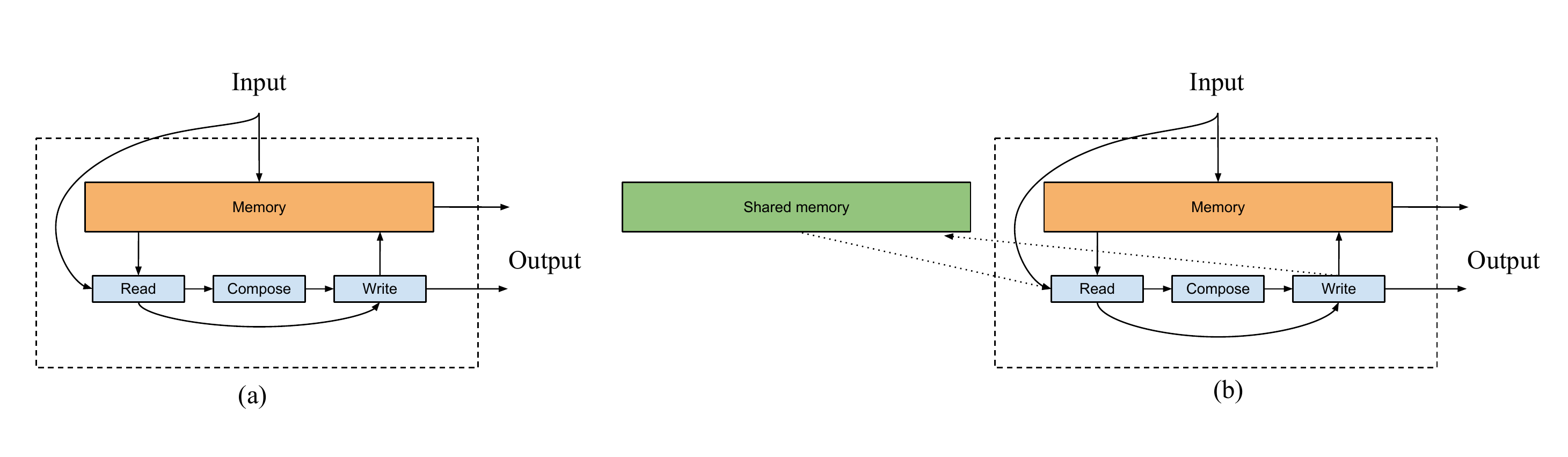}
        \vspace*{-1.0cm}  
        \caption{\label{fig:skip-gram} High-level architectures of the Neural Semantic Encoders. NSE reads and writes its own encoding memory in each time step (a). MMA-NSE accesses multiple relevant memories simultaneously (b).}
        \label{figure:NSE}
\end{figure*}

One of the pioneering work that attempts to extend deep neural networks with an external memory is Neural Turing Machines (NTM) \cite{graves2014neural}. NTM implements a centralized controller and a fixed-sized random access memory. The NTM memory is addressable by both content (i.e. soft attention) and location based access mechanisms. The authors evaluated NTM on algorithmic tasks such as copying and sorting sequences.

\textbf{Comparison with Neural Turing Machines:} NSE addresses certain drawbacks of NTM. NTM has a single centralized controller, which is usually an MLP or RNN while NSE takes a modular approach. The main controller in NSE is decomposed into three separate modules, each of which performs for \textit{read}, \textit{compose} or \textit{write} operation. In NSE, the \textit{compose} module is introduced in addition to the standard memory update operations (i.e. read-write) in order to process the memory entries and input information.

The main advantage of NSE over NTM is in its memory update. Despite its sophisticated addressing mechanism, the NTM controller does not have mechanism to avoid \textit{information collision} in the memory. Particularly the NTM controller emits two separate set of access weights (i.e. read weight and erase and write weights) that do not explicitly encode the knowledge about where information is read from and written to. Moreover the fixed-size memory in NTM has no memory allocation or de-allocation protocol. Therefore unless the controller is intelligent enough to track the previous read/write information, which is hard for an RNN when processing long sequences, the memory content is overlapped and information is overwritten throughout different time scales. We think that this is a potential reason that makes NTM hard to train and makes the training not stable. We also note that the effectiveness of the location based addressing introduced in NTM is unclear. 
In NSE, we introduce a novel and systematic memory update approach based on the soft attention mechanism. NSE writes new information to the most recently read memory locations. This is accomplished by sharing the same memory key vector between the read and write modules. The NSE memory update is scalable and potentially more robust to train. NSE is provided with a variable sized memory and thus unlike NTM, the size of the NSE memory is more relaxed. The novel memory update mechanism and the variable sized memory together prevent NSE from the \textit{information collision} issue and avoid the need of the memory allocation or de-allocation protocols.
Each memory location of the NSE memory stores a token representation in input sequence during encoding. This provides NSE with an anytime-access to the entire input sequence including the tokens from the future time scales, which is not permitted in NTM, RNN and attention-based encoders.

Lastly, NTM addresses small algorithmic problems while NSE focuses on a set of large-scale language understanding tasks.

The RNNSearch model proposed in \cite{bahdanau:15} can be seen as a variation of memory augmented networks due to its ability to read the historic output states of RNNs with soft attention. The work of \cite{sukhbaatar2015end} combines the soft attention with Memory Networks (MemNNs) \cite{weston:15}. Similar to RNNSearch, MemNNs are designed with non-writable memories. It constructs layered memory representations and showed promising results on both artificial and real question answering tasks. We note that RNNSearch and MemNNs avoid the memory update and management overhead by simply using a non-writable memory storage. Another variation of MemNNs is Dynamic Memory Network \cite{ankit16} that is equipped with an episodic memory and seems to be flexible in different settings.

Although NSE differs from other memory-augumented NN models in many aspects, they all use soft attention mechanism with a type of similarity measures to retrieve relevant information from the external memory. For example, NTM implements cosine similarity and MemNNs use vector dot product. NSE uses the vector dot product for the similarity measure in NSE because it is faster to compute.


Other related work includes Neural Program-Interpreters \cite{reed2015neural}, which learns to run sub-programs and to compose them for high-level programs. It uses execution traces to provide the full supervision. Researchers have also explored ways to add unbounded memory to LSTM \cite{grefenstette2015learning} using a particular data structure. Although this type of architecture provides a flexible capacity to store information, the memory access is constrained by the data structure used for the memory bank, such as stack and queue.

Overall it is expensive to train and to scale the previously proposed memory-based models. Most models required a set of clever engineering tricks to work successfully. 
Most of the aforementioned memory augmented neural networks have been tested on synthetic tasks whereas in this paper we evaluated NSE on a wide range of real and large-scale natural language applications.

\section{Proposed Approach}

Our training set consists of $N$ examples $\lbrace X^i,Y^i \rbrace^N_{i=1}$, where the input $X^i$ is a sequence $w^i_1, w^i_2, \ldots, w^i_{T_i}$ of tokens while the output $Y^i$ can be either a single target or a sequence. We transform each input token $w_t$ to its word embedding $x_t$.

Our Neural Semantic Encoders (NSE) model has four main components: read, compose and write modules and an encoding memory $M \in R^{k \times l}$ with a variable number of slots, where $k$ is the embedding dimension and $l$ is the length of the input sequence. Each memory slot vector $m_t \in R^{k}$ corresponds to the vector representation of information about word $w_t$ in memory. In particular, the memory is initialized by the embedding vectors $\lbrace x_t \rbrace^l_{t=1}$ and is evolved over time, through \textit{read}, \textit{compose} and \textit{write} operations. Figure $~\ref{figure:NSE}$ (a) illustrates the architecture of NSE.

\subsection{Read, Compose and Write}

NSE performs three main operations in every time step. After initializing the memory slots with the corresponding input representations, NSE processes an embedding vector $x_t$ and retrieves a memory slot $m_{r,t}$ that is expected to be associatively coherent (i.e. semantically associated) with the current input word $w_t$.\footnote{Such a coherence is calculated by a soft attention with dot product similarity.} The slot location $r$ (ranging from 1 to $l$) is defined by a key vector $z_t$ which the read module emits by attending over the memory slots. The compose module implements a composition operation that combines the memory slot with the current input. The write module then transforms the composition output to the encoding memory space and writes the resulting new representation into the slot location of the memory. Instead of composing the raw embedding vector $x_t$, we use the hidden state $o_t$ produced by the read module at time $t$

Concretely, let $e_l \in R^l$ and $e_k \in R^k$ be vectors of ones and given a read function $f^{LSTM}_r$, a composition $f^{MLP}_c$ and a write $f^{LSTM}_w$ NSE in Figure $~\ref{figure:NSE}$ (a) computes the key vector $z_t$, the output state $h_t$, and the encoding memory $M_t$ in time step $t$ as

\begin{equation}
o_t = f^{LSTM}_r(x_t)
\end{equation}
\begin{equation}
z_t = softmax(o_t^{\top} M_{t-1})
\end{equation}
\begin{equation}
m_{r,t} = z_t^{\top} M_{t-1}
\end{equation}
\begin{equation}
c_t = f^{MLP}_c(o_t, m_{r,t})
\end{equation}
\begin{equation}
h_t = f^{LSTM}_w(c_t)
\end{equation}
\begin{equation}
M_t = M_{t-1}(\textbf{1}-(z_t \otimes e_k)^{\top}) + (h_t \otimes e_l) (z_t \otimes e_k)^{\top}
\end{equation}
where $\textbf{1}$ is a matrix of ones, $\otimes$ denotes the outer product which duplicates its left vector $l$ or $k$ times to form a matrix. The read function $f^{LSTM}_r$ sequentially maps the word embeddings to the internal space of the memory $M_{t-1}$. Then Equation 2 looks for the slots related to the input by computing association degree between each memory slot and the hidden state $o_t$. We calculate the association degree by the dot product and transform this scores to the fuzzy key vector $z_t$ by normalizing with $softmax$ function. Since our key vector is fuzzy, the slot to be composed is retrieved by taking weighted sum of the all slots as in Equation 3. This process can also be seen as the soft attention mechanism \cite{bahdanau:15}. In Equation 4 and 5, we compose and process the retrieved slot with the current hidden state and map the resulting vector to the encoder output space. Finally, we write the new representation to the memory location pointed by the key vector in Equation 6 where the key vector $z_t$ emitted by the read module is reused to inform the write module of the most recently read slots. First the slot information that was retrieved is erased and then the new representation is located. NSE performs this iterative process until all words in the input sequence are read. The encoding memories $\lbrace M \rbrace ^T_{t=1}$ and output states $\lbrace h \rbrace ^T_{t=1}$ are further used for the tasks.

Although NSE reads a single word at a time, it has an anytime-access to the entire sequence stored in the encoding memory. With the encoding memory, NSE maintains a mental image of the input sequence. The memory is initialized with the raw embedding vector at time $t=0$. We term such a freshly initialized memory a \textit{baby} memory. As NSE reads more input content in time, the \textit{baby} memory evolves and refines the encoded mental image. 

The read $f^{LSTM}_r$, the composition $f^{MLP}_c$ and the write $f^{LSTM}_w$ functions are neural networks and are the training parameters in our NSE. As the name suggests, we use LSTM and multi-layer perceptron (MLP) in this paper. Since NSE is fully differentiable, it can be trained with any gradient descent optimizer.  

\subsection{Shared and Multiple Memory Accesses}

For sequence to sequence transduction tasks like question answering, natural language inference and machine translation, it is beneficial to access other relevant memories in addition to its own one. The shared or the multiple memory access allows a set of NSEs to exchange knowledge representations and to communicate with each other to accomplish a particular task throughout the encoding memory. 

NSE can be extended easily, so that it is able to read from and write to multiple memories simultaneously or multiple NSEs are able to access a shared memory. Figure $~\ref{figure:NSE}$ (b) depicts a high-level architectural diagram of a multiple memory access-NSE (MMA-NSE). The first memory (in green) is the shared memory accessed by more than one NSEs. Given a shared memory $M^{n} \in R^{k \times n}$ that has been encoded by processing a relevant sequence with length $n$, MMA-NSE with the access to one relevant memory is defined as

\begin{equation}
o_t = f^{LSTM}_r(x_t)
\end{equation}
\begin{equation}
z_t = softmax(o_t^{\top} M_{t-1})
\end{equation}
\begin{equation}
m_{r,t} = z_t^{\top} M_{t-1}
\end{equation}
\begin{equation}
z^n_t = softmax(o_t^{\top} M^n_{t-1})
\end{equation}
\begin{equation}
m^n_{r,t} = {z^n}_t^{\top} M^n_{t-1}
\end{equation}
\begin{equation}
c_t = f^{MLP}_c(o_t, m_{r,t}, m^n_{r,t})
\end{equation}
\begin{equation}
h_t = f^{LSTM}_w(c_t)
\end{equation}
\begin{equation}
M_t = M_{t-1}(\textbf{1}-(z_t \otimes e_k)^{\top}) + (h_t \otimes e_l) (z_t \otimes e_k)^{\top}
\end{equation}
\begin{equation}
M^n_t = M^n_{t-1}(\textbf{1}-(z^n_t \otimes e_k)^{\top}) + (h_t \otimes e_n) (z^n_t \otimes e_k)^{\top}
\end{equation}
and this is almost the same as standard NSE. The read module now emits the additional key vector $z^n_t$ for the shared memory and the composition function $f^{MLP}_c$ combines more than one slots.

In MMA-NSE, the different memory slots are retrieved from the shared memories depending on their encoded semantic representations. They are then composed together with the current input and written back to their corresponding slots. Note that MMA-NSE is capable of accessing a variable number of relevant shared memories once a composition function that takes in dynamic inputs is chosen.




\section{Experiments}

\begin{table*}[t]
\begin{center}
\small
\begin{tabular}{c|c|c|c|c}
\hline 
Model & $d$ & $|\theta|_M$ & Train	& Test \\
\hline
\multicolumn{1}{l|}{Classifier with handcrafted features \cite{bowman:15}} & \multicolumn{1}{|r|}{-} & \multicolumn{1}{|r|}{-} & \multicolumn{1}{|r|}{99.7} & \multicolumn{1}{|r}{78.2} \\
\hline
\multicolumn{1}{l|}{LSTM encoders \cite{bowman:15}} & \multicolumn{1}{|r|}{300} & \multicolumn{1}{|r|}{3.0M} & \multicolumn{1}{|r|}{83.9} & \multicolumn{1}{|r}{80.6} \\
\multicolumn{1}{l|}{Dependency Tree CNN encoders \cite{Lili16}} & \multicolumn{1}{|r|}{300} & \multicolumn{1}{|r|}{3.5M} & \multicolumn{1}{|r|}{83.3} & \multicolumn{1}{|r}{82.1} \\
\multicolumn{1}{l|}{SPINN-PI encoders \cite{BowmanGRGMP16}} & \multicolumn{1}{|r|}{300} & \multicolumn{1}{|r|}{3.7M} & \multicolumn{1}{|r|}{89.2} & \multicolumn{1}{|r}{83.2} \\
\multicolumn{1}{l|}{NSE} & \multicolumn{1}{|r|}{300} & \multicolumn{1}{|r|}{3.4M} & \multicolumn{1}{|r|}{86.2} & \multicolumn{1}{|r}{84.6} \\ 
\multicolumn{1}{l|}{MMA-NSE} & \multicolumn{1}{|r|}{300} & \multicolumn{1}{|r|}{6.3M} & \multicolumn{1}{|r|}{87.1} & \multicolumn{1}{|r}{\textbf{84.8}} \\ 
\hline
\multicolumn{1}{l|}{LSTM attention \cite{rocktaschel:16}} & \multicolumn{1}{|r|}{100} & \multicolumn{1}{|r|}{242K} & \multicolumn{1}{|r|}{85.4} & \multicolumn{1}{|r}{82.3} \\
\multicolumn{1}{l|}{LSTM word-by-word attention \cite{rocktaschel:16}} & \multicolumn{1}{|r|}{100} & \multicolumn{1}{|r|}{252K} & \multicolumn{1}{|r|}{85.3} & \multicolumn{1}{|r}{83.5} \\
\multicolumn{1}{l|}{MMA-NSE attention} & \multicolumn{1}{|r|}{300} & \multicolumn{1}{|r|}{6.5M} & \multicolumn{1}{|r|}{86.9} & \multicolumn{1}{|r}{85.4} \\ 
\multicolumn{1}{l|}{mLSTM word-by-word attention \cite{WangJ15b}} & \multicolumn{1}{|r|}{300} & \multicolumn{1}{|r|}{1.9M} & \multicolumn{1}{|r|}{92.0} & \multicolumn{1}{|r}{86.1} \\
\multicolumn{1}{l|}{LSTMN with deep attention fusion \cite{ChengDL16}} & \multicolumn{1}{|r|}{450} & \multicolumn{1}{|r|}{3.4M} & \multicolumn{1}{|r|}{89.5} & \multicolumn{1}{|r}{86.3} \\
\multicolumn{1}{l|}{Decomposable attention model \cite{parikh2016decomposable}} & \multicolumn{1}{|r|}{200} & \multicolumn{1}{|r|}{582K} & \multicolumn{1}{|r|}{90.5} & \multicolumn{1}{|r}{86.8} \\
\multicolumn{1}{l|}{Full tree matching NTI-SLSTM-LSTM global attention \cite{munkhdalai2016neural}} & \multicolumn{1}{|r|}{300} & \multicolumn{1}{|r|}{3.2M} & \multicolumn{1}{|r|}{88.5} & \multicolumn{1}{|r}{\textbf{87.3}} \\
\hline
\end{tabular}
\end{center}
\caption{\label{table:snli}Training and test accuracy on natural language inference task. $d$ is the word embedding size and $|\theta|_M$ the number of model parameters.}
\end{table*}

We describe in this section experiments on five different tasks, in order to show that NSE can be effective and flexible in different settings.\footnote{Code for the experiments and NSEs is available at https://bitbucket.org/tsendeemts/nse.} We report results on natural language inference, question answering (QA), sentence classification, document sentiment analysis and machine translation. All five tasks challenge a model in terms of language understanding and semantic reasoning. 

The models are trained using Adam \cite{kingma:14} with hyperparameters selected on development set. We chose two one-layer LSTM for read/write modules on the tasks other than QA on which we used two-layer LSTM. The pre-trained 300-D Glove 840B vectors and 100-D Glove 6B vectors \cite{pennington:14} were obtained for the word embeddings.\footnote{http://nlp.stanford.edu/projects/glove/} The word embeddings are fixed during training. The embeddings for out-of-vocabulary words were set to zero vector. We crop or pad the input sequence to a fixed length. A padding vector was inserted when padding. The models were regularized by using dropouts and an $l_2$ weight decay.\footnote{More detail on hyper-parameters can be found in code.}

\subsection{Natural Language Inference}

The natural language inference is one of the main tasks in language understanding. This task tests the ability of a model to reason about the semantic relationship between two sentences. In order to perform well on the task, NSE should be able to capture sentence semantics and be able to reason the relation between a sentence pair, i.e., whether a premise-hypothesis pair is entailing, contradictory or neutral. We conducted experiments on the Stanford Natural Language Inference (SNLI) dataset \cite{bowman:15}, which consists of 549,367/9,842/9,824 premise-hypothesis pairs for train/dev/test sets and target label indicating their relation. 

Following the setting in \cite{Lili16,BowmanGRGMP16} the NSE output for each sentence was the input to a MLP, where the input layer computes the concatenation $[h^p_l;h^h_l]$, absolute difference $h^p_l-h^h_l$ and elementwise product $h^p_l \cdot h^h_l$ of the two sentence representations. In addition, the MLP has a hidden layer with 1024 units with $ReLU$ activation and a $softmax$ layer. We set the batch size to 128, the initial learning rate to 3e-4 and $l_2$ regularizer strength to 3e-5, and train each model for 40 epochs. The write/read neural nets and the last linear layer were regularized by using 30\% dropouts.

We evaluated three different variations of NSE show in Table \ref{table:snli}. The NSE model encodes each sentence simultaneously by using a separate memory for each sentence. The second model - MMA-NSE first encodes the premise and then the hypothesis sentence by sharing the premise encoded memory in addition to the hypothesis memory. For the third model, we use inter-sentence attention which selectively reconstructs the premise representation.

Table \ref{table:snli} shows the results of our models along with the results of published methods for the task. The classifier with handcrafted features extracts a set of lexical features. The next group of models are based on sentence encoding. While most of the sentence encoder models rely solely on word embeddings, the dependency tree CNN and the SPINN-PI models make use of sentence parser output. The SPINN-PI model is similar to NSE in spirit that it also explicitly computes word composition. However, the composition in the SPINN-PI is guided by supervisions from a dependency parser. NSE outperformed the previous sentence encoders on this task. The MMA-SNE further slightly improved the result, indicating that reading the premise memory is helpful while encoding the hypothesis. 

The last set of methods designs inter-sentence relation with parameterized soft attention \cite{bahdanau:15}. Our MMA-NSE attention model is similar to the LSTM attention model. Particularly, it attends over the premise encoder outputs  $\lbrace h^p \rbrace ^T_{t=1}$ in respect to the final hypothesis representation $h^h_l$ and constructs an attentively blended vector of the premise. This model obtained 85.4\% accuracy score. The best performing model for this task performs tree matching with attention mechanism and LSTM.

\subsection{Answer Sentence Selection}

\begin{table}[b]
\begin{center}
\small
\begin{tabular}{c|c|c}
\hline 
Model & MAP & MRR \\
\hline
\multicolumn{1}{l|}{Classifier with features \cite{yih13}} & \multicolumn{1}{|r|}{0.5993} & \multicolumn{1}{|r}{0.6068} \\
\hline
\multicolumn{1}{l|}{Paragraph Vector \cite{le2014}} & \multicolumn{1}{|r|}{0.5110} & \multicolumn{1}{|r}{0.5160} \\
\multicolumn{1}{l|}{Bigram-CNN \cite{yu2014}} & \multicolumn{1}{|r|}{0.6190} & \multicolumn{1}{|r}{0.6281} \\
\multicolumn{1}{l|}{3-layer LSTM \cite{miao2016}} & \multicolumn{1}{|r|}{0.6552} & \multicolumn{1}{|r}{0.6747} \\
\multicolumn{1}{l|}{3-layer LSTM attention \cite{miao2016}} & \multicolumn{1}{|r|}{0.6639} & \multicolumn{1}{|r}{0.6828} \\
\multicolumn{1}{l|}{NASM \cite{miao2016}} & \multicolumn{1}{|r|}{0.6705} & \multicolumn{1}{|r}{0.6914} \\
\multicolumn{1}{l|}{MMA-NSE attention} & \multicolumn{1}{|r|}{\bf 0.6811} & \multicolumn{1}{|r}{\bf 0.6993} \\
\hline
\end{tabular}
\end{center}
\caption{\label{table:qa}Experiment results on answer sentence selection.}
\end{table}

Answer sentence selection is an integral part of the open-domain question answering. For this task, a model is trained to identify the correct sentences that answer a factual question, from a set of candidate sentences. We experiment on WikiQA dataset constructed from Wikipedia \cite{yang2015wikiqa}. The dataset contains 20,360/2,733/6,165 QA pairs for train/dev/test sets. 

The MLP setup used in the language inference task is kept same, except that we now replace the $softmax$ layer with a $sigmoid$ layer and model the following conditional probability distribution.
\begin{equation}
p_{\theta}(y=1|h^q_l,h^a_l) = sigmoid(o^{QA})
\end{equation}
where $h^q_l$ and $h^a_l$ are the question and the answer encoded vectors and $o^{QA}$ denotes the output of the hidden layer of the MLP. We trained the MMA-NSE attention model to minimize the $sigmoid$ cross entropy loss. MMA-NSE first encodes the answers and then the questions by accessing its own and the answer encoding memories. In our preliminary experiment, we found that the multiple memory access and the attention over answer encoder outputs  $\lbrace h^a \rbrace ^T_{t=1}$ are crucial to this problem. Following previous work, we adopt MAP and MRR as the evaluation metrics for this task.\footnote{We used \textit{trec\_eval} script to calculate the evaluation metrics}

We set the batch size to 4 and the initial learning rate to 1e-5, and train the model for 10 epochs. We used 40\% dropouts after word embeddings and no $l_2$ weight decay. The word embeddings are pre-trained 300-D Glove 840B vectors. For this task, a linear mapping layer transforms the 300-D word embeddings to the 512-D LSTM inputs.

Table \ref{table:qa} presents the results of our model and the previous models for the task.\footnote{Inclusion of simple word count feature improves the performance by around 0.15-0.3 across the board} The classifier with handcrafted features is a SVM model trained with a set of features. The Bigram-CNN model is a simple convolutional neural net. While the LSTM and LSTM attention models outperform the previous best result by nearly 5-6\% by implementing deep LSTM with three hidden layers, NASM improves it further and sets a strong baseline by combining variational auto-encoder \cite{kingma2014auto} with the soft attention. Our MMA-NSE attention model exceeds the NASM by approximately 1\% on MAP and 0.8\% on MRR for this task.

\subsection{Sentence Classification}

We evaluated NSE on the Stanford Sentiment Treebank (SST) \cite{socher2013recursive}. This dataset comes with standard train/dev/test sets and two subtasks: binary sentence classification or fine-grained classification of five classes. We trained our model on the text spans corresponding to labeled phrases in the training set and evaluated the model on the full sentences. 

The sentence representations were passed to a two-layer MLP for classification. The first layer of the MLP has $ReLU$ activation and 1024 or 300 units for binary or fine-grained setting. The second layer is a $softmax$ layer. The read/write modules are two one-layer LSTM with 300 hidden units and the word embeddings are the pre-trained 300-D Glove 840B vectors. We set the batch size to 64, the initial learning rate to 3e-4 and $l_2$ regularizer strength to 3e-5, and train each model for 25 epochs. The write/read neural nets and the last linear layer were regularized by 50\% dropouts. 

Table \ref{table:sent} compares the result of our model with the state-of-the-art methods on the two subtasks. Most best performing methods exploited the parse tree provided in the treebank on this task with the exception of the DMN. The Dynamic Memory Network (DMN) model is a memory-augmented network. Our model outperformed the DMN and set the state-of-the-art results on both subtasks.

\begin{table}[t]
\begin{center}
\small
\begin{tabular}{c|c|c}
\hline 
Model & Bin & FG \\
\hline
\multicolumn{1}{l|}{RNTN \cite{socher2013recursive}} & \multicolumn{1}{|r|}{85.4} & \multicolumn{1}{|r}{45.7} \\
\multicolumn{1}{l|}{Paragraph Vector \cite{le2014}} & \multicolumn{1}{|r|}{87.8} & \multicolumn{1}{|r}{48.7} \\
\multicolumn{1}{l|}{CNN-MC \cite{kim:13}} & \multicolumn{1}{|r|}{88.1} & \multicolumn{1}{|r}{47.4} \\
\multicolumn{1}{l|}{DRNN \cite{irsoy15}} & \multicolumn{1}{|r|}{86.6} & \multicolumn{1}{|r}{49.8} \\
\multicolumn{1}{l|}{2-layer LSTM\cite{tai2015improved}} & \multicolumn{1}{|r|}{86.3} & \multicolumn{1}{|r}{46.0} \\
\multicolumn{1}{l|}{Bi-LSTM\cite{tai2015improved}} & \multicolumn{1}{|r|}{87.5} & \multicolumn{1}{|r}{49.1} \\
\multicolumn{1}{l|}{CT-LSTM\cite{tai2015improved}} & \multicolumn{1}{|r|}{88.0} & \multicolumn{1}{|r}{51.0} \\
\multicolumn{1}{l|}{DMN \cite{ankit16}} & \multicolumn{1}{|r|}{88.6} & \multicolumn{1}{|r}{52.1} \\
\multicolumn{1}{l|}{NSE} & \multicolumn{1}{|r|}{\bf 89.7} & \multicolumn{1}{|r}{\bf 52.8} \\
\hline
\end{tabular}
\end{center}
\caption{\label{table:sent}Test accuracy for sentence classification. Bin: Binary, FG: fine-grained 5 classes.}
\end{table}

\subsection{Document Sentiment Analysis}

We evaluated our models for document-level sentiment analysis on two publically available large-scale datasets: the IMDB  consisting of 335,018 movie reviews and 10 different classes and Yelp 13 consisting of 348,415 restaurant reviews and 5 different classes. Each document in the datasets is associated with human ratings and we used these ratings as gold labels for sentiment classification. Particularly, we used the pre-split datasets of \cite{tang:15}.

We stack a NSE or LSTM on the top of another NSE for document modeling. The first NSE encodes the sentences and the second NSE or LSTM takes sentence encoded outputs and constructs document representations. The document representation is given to a output $softmax$ layer. The whole network is trained jointly by backpropagating the cross entropy loss. We used one-layer LSTM with 100 hidden units for the read/write modules and the pre-trained 100-D Glove 6B vectors for this task. We set the batch size to 32, the initial learning rate to 3e-4 and $l_2$ regularizer strength to 1e-5, and trained each model for 50 epochs. The write/read neural nets and the document-level NSE/LSTM were regularized by 15\% dropouts and the softmax layer by 20\% dropouts. In order to speedup the training, we created document buckets by considering the number of sentences per document, i.e., documents with the same number of sentences were put together in the same bucket. The buckets were shuffled and updated per epoch. We did not use curriculum scheduling \cite{bengio2009curriculum}, although it is observed to help sequence training.

\begin{table}[tp]
\begin{center}
\small
\begin{tabular}{c|c|c|c|c}
\hline
\multirow{2}{*}{Model} & \multicolumn{2}{|c|}{Yelp 13} & \multicolumn{2}{|c}{IMDB} \\
\hhline{~----} & Acc & MSE & Acc & MSE\\
\hline

\multicolumn{1}{l|}{Classifier \cite{tang:15}} & \multicolumn{1}{|r|}{59.8} & \multicolumn{1}{|r|}{0.68} & \multicolumn{1}{|r|}{40.5} & \multicolumn{1}{|r}{3.56} \\

\hline

\multicolumn{1}{l|}{PV \cite{tang:15}} & \multicolumn{1}{|r|}{57.7} & \multicolumn{1}{|r|}{0.86} & \multicolumn{1}{|r|}{34.1} & \multicolumn{1}{|r}{4.69} \\
\multicolumn{1}{l|}{CNN \cite{tang:15}} & \multicolumn{1}{|r|}{59.7} & \multicolumn{1}{|r|}{0.76} & \multicolumn{1}{|r|}{37.6} & \multicolumn{1}{|r}{3.30} \\
\multicolumn{1}{l|}{Conv-GRNN \cite{tang:15}} & \multicolumn{1}{|r|}{63.7} & \multicolumn{1}{|r|}{0.56} & \multicolumn{1}{|r|}{42.5} & \multicolumn{1}{|r}{2.71} \\
\multicolumn{1}{l|}{LSTM-GRNN \cite{tang:15}} & \multicolumn{1}{|r|}{65.1} & \multicolumn{1}{|r|}{0.50} & \multicolumn{1}{|r|}{45.3} & \multicolumn{1}{|r}{3.00} \\
\multicolumn{1}{l|}{NSE-NSE} & \multicolumn{1}{|r|}{66.6} & \multicolumn{1}{|r|}{0.48} & \multicolumn{1}{|r|}{\bf 48.3} & \multicolumn{1}{|r}{\textbf{\bf 1.94}} \\
\multicolumn{1}{l|}{NSE-LSTM} & \multicolumn{1}{|r|}{\bf 67.0} & \multicolumn{1}{|r|}{\bf 0.47} & \multicolumn{1}{|r|}{48.1} & \multicolumn{1}{|r}{1.98} \\

\hline
\end{tabular} 
\end{center}
\caption{\label{table:doc}Results of document-level sentiment classification. PV: paragraph vector, Acc: accuracy, and MSE: mean squared error.}
\end{table}

Table \ref{table:doc} shows our results. We report two performance metrics: accuracy and MSE. The best results on the task were previously obtained by Conv-GRNN and LSTM-GRNN, which are also stacked models. These models first learn the sentence representations with a CNN or LSTM and then combine them for document representation using a gated recurrent neural network (GRNN). Our NSE models outperformed the previous state-of-the-art models in terms of both accuracy and MSE, by approximately 2-3\%. On the other hand, all systems tend to show poor results on the IMDB dataset. That is, the IMDB dataset contains longer documents than the Yelp 13 and it has 10 classes while the Yelp 13 dataset has five classes to distinguish.\footnote{The average number of sentences and words in a document for IMDB: 14, 152 and Yelp 13: 9, 326} The stacked NSEs (NSE-NSE) performed slightly better than the NSE-LSTM on the IMDB dataset. This is possibly due to the encoding memory of the document level NSE that preserves the long dependency in documents with a large number of sentences.

\begin{table}[tp]
\begin{center}
\small
\begin{tabular}{c|c|c|c}
\hline 
Model & Train & Dev & Test \\
\hline
\multicolumn{1}{l|}{Baseline LSTM-LSTM} & \multicolumn{1}{|r|}{28.06} & \multicolumn{1}{|r|}{17.96} & \multicolumn{1}{|r}{17.02} \\
\multicolumn{1}{l|}{NSE-LSTM} & \multicolumn{1}{|r|}{28.73} & \multicolumn{1}{|r|}{17.67} & \multicolumn{1}{|r}{17.13} \\
\multicolumn{1}{l|}{NSE-NSE} & \multicolumn{1}{|r|}{29.89} & \multicolumn{1}{|r|}{\bf 18.53} & \multicolumn{1}{|r}{\bf 17.93} \\
\hline
\end{tabular}
\end{center}
\caption{\label{table:nmt}BLEU scores for English-German translation task.}
\end{table}

\subsection{Machine Translation}

Lastly, we conducted an experiment on neural machine translation (NMT). The NMT problem is mostly defined within the encoder-decoder framework \cite{kalchbrenner2013recurrent,cho2014learning,sutskever2014sequence}. 
The encoder provides the semantic and syntactic information about the source sentences to the decoder and the decoder generates the target sentences by conditioning on this information and its partially produced translation. 
For an efficient encoding, the attention-based NTM was introduced \cite{bahdanau:15}.

For NTM, we implemented three different models. The first model is a baseline model and is similar to the one proposed in \cite{bahdanau:15} (RNNSearch). This model (LSTM-LSTM) has two LSTM for the encoder/decoder and has the soft attention neural net, which attends over the source sentence and constructs a focused encoding vector for each target word. The second model is an NSE-LSTM encoder-decoder which encodes the source sentence with NSE and generates the targets with the LSTM network by using the NSE output states and the attention network. The last model is an NSE-NSE setup, where the encoding part is the same as the NSE-LSTM while the decoder NSE now uses the output state and has an access to the encoder memory, i.e., the encoder and the decoder NSEs access a shared memory. The memory is encoded by the first NSEs and then read/written by the decoder NSEs. 
We used the English-German translation corpus from the IWSLT 2014 evaluation campaign \cite{cettoloEtAl:EAMT2012}. The corpus consists of sentence-aligned translation of TED talks. The data was pre-processed and lowercased with the Moses toolkit.\footnote{https://github.com/moses-smt/mosesdecoder} We merged the dev2010 and dev2012 sets for development and the tst2010, tst2011 and tst2012 sets for test data\footnote{We modified \textit{prepareData.sh} script: https://github.com/facebookresearch/MIXER}. Sentence pairs with length longer than 25 words were filtered out. This resulted in 110,439/4,998/4,793 pairs for train/dev/test sets. We kept the most frequent 25,000 words for the German dictionary. The English dictionary has 51,821 words. The 300-D Glove 840B vectors were used for embedding the words in the source sentence whereas a lookup embedding layer was used for the target German words. Note that the word embeddings are usually optimized along with the NMT models. However, for the evaluation purpose we in this experiment do not optimize the English word embeddings. Besides, we do not use a beam search to generate the target sentences.

\begin{figure*}[tp]
    \centering \includegraphics[width=1\textwidth]{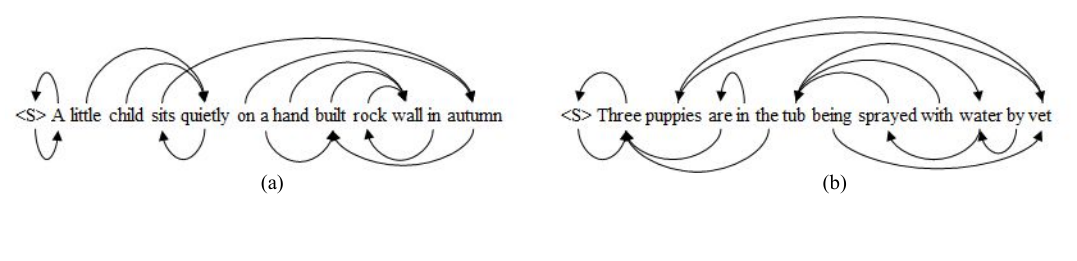}
       \vspace*{-1.0cm}  
        \caption{\label{fig:asoc}
        Word association or composition graphs produced by NSE memory access. The directed arcs connect the words that are composed via \textit{compose} module. The source nodes are input words and the destination nodes (pointed by the arrows) correspond to the accessed memory slots.  $<S>$ denotes the beginning of sequence.}
\end{figure*}

The LSTM encoder/decoders have two layers with 300 units. The NSE read/write modules are two one-layer LSTM with the same number of units as the LSTM encoder/decoders. This ensures that the number of parameters of the models is roughly the equal. The models were trained to minimize word-level cross entropy loss and were regularized by 20\% input dropouts and the 30\% output dropouts. We set the batch size to 128, the initial learning rate to 1e-3 for LSTM-LSTM and 3e-4 for the other models and $l_2$ regularizer strength to 3e-5, and train each model for 40 epochs. We report BLEU score for each models.\footnote{We computed the BLEU score with \textit{multi-bleu.perl} script of the Moses toolkit}

Table \ref{table:nmt} reports our results. The baseline LSTM-LSTM encoder-decoder (with attention) obtained 17.02 BLEU on the test set. The NSE-LSTM improved the baseline slightly. Given this very small improvement of the NSE-LSTM, it is unclear whether the NSE encoder is helpful in NMT. However, if we replace the LSTM decoder with another NSE and introduce the shared memory access to the encoder-decoder model (NSE-NSE), we improve the baseline result by almost 1.0 BLEU. The NSE-NSE model also yields an increasing BLEU score on dev set. 
The result demonstrates that the attention-based NMT systems can be improved by a shared-memory encoder-decoder model. In addition, memory-based NMT systems should perform well on translation of long sequences by preserving long term dependencies.

\section{Qualitative Analysis}
\subsection{Memory Access and Compositionality}
NSE is capabable of performing multiscale composition by retrieving associative slots for a particular input at a time step. We analyzed the memory access order and the compositionality of memory slot and the input word in the NSE model trained on the SNLI data.

Figure \ref{fig:asoc} shows the word association graphs for the two sentence picked from SNLI test set. The association graph was constructed by inspecting the key vector $z$. For an input word, we connect it to the most active slot pointed by $z$\footnote{Since $z$ is fuzzy, we visualize the highest scoring slot. For a few inputs, $z$ pointed to a slot corresponding to the same word. In this case, we masked out those slots and showed the second best scoring slot.}. 

Note the graph components clustered around the semantically rich words: \textit{"sits"}, \textit{"wall"} and \textit{"autumn"} (a) and \textit{"Three"}, \textit{"puppies"}, \textit{"tub"} and \textit{"vet"} (b). The memory slots corresponding to words that are semantically rich in the current context are the most frequently accessed. The graph is able to capture certain syntactic structures including phrases (e.g., "hand built rock wall") and modifier relations (between "sits" and "quietly" and between "tub" and "sprayed with water"). Another interesting property is that the model tends to perform sensible compositions while processing the input sentence. For example, NSE retrieved the memory slot corresponding to \textit{"wall"} or \textit{"Three"} when reading the input \textit{"rock"} or \textit{"are"}.

In Appendix \ref{sup:mem}, we show a step-by-step visualization of NSE memory states for the first sentence. Note how the encoding memory is evolved over time. In time step four ($t=4$), the memory slot for \textit{"quietly"} encodes information about \textit{"quiet(ly) little child"}. When $t=6$, the model forms another composition involving \textit{"quietly"}, \textit{"quietly sits"}. In the last time step, we are able to find the most or the least frequently accessed slots in the memory. The least accessed slots correspond to function words while the frequently accessed slots are content words and tend to carry out rich semantics and intrinsic compositions found in the input sentence. Overall the model is less constrained and is able to compose multiword expressions. 

\section{Conclusion}

Our proposed memory augmented neural networks have achieved the state-of-the-art results when evaluated on five representative NLP tasks. 
NSE is capable of building an efficient architecture of the single, shared and multiple memory accesses for a specific NLP task. 
For example, for the NLI task NSE accesses premise encoded memory when processing hypothesis. For the QA task, NSE accesses answer encoded memory when reading question for QA. In machine translation, NSE shares a single encoded memory between encoder and decoder. Such flexibility in the architectural choice of the NSE memory access allows for the robust models for a better performance. 

The initial state of the NSE memory stores information about each word in the input sequence. We in this paper used word embeddings to represent the words in the memory. Different variations of word representations such as character-based models are left to be evaluated for memory initialization in the future. We plan to extend NSE so that it learns to select and access a relevant subset from a memory set.
One could also explore unsupervised variations of NSE, for example, to train them to produce encoding memory and representation vector of entire sentences or documents using either new or existing models such as the skip-gram model \cite{mikolov:13}.

\section*{Acknowledgments}

We would like to thank Abhyuday Jagannatha and the anonymous reviewers for
their insightful comments and suggestions.
This work was supported in part by the grant HL125089 from the National Institutes of Health (NIH). Any opinions, findings and conclusions or recommendations expressed in this material are those of the authors and do not necessarily 
reflect those of the sponsor.

\begingroup
	\small
	\setlength{\bibsep}{0pt plus 0.3ex}
	\bibliography{nips_2016}
	\bibliographystyle{unsrt}
\endgroup

\newpage

\appendix
\section{Step-by-step visualization of memory states in NSE}
\label{sup:mem}
Each small table represents the memory state at a single time step. The current time step and input token are listed on the top of the table. The memory slots pointed by the query vector is highlighted in red color. The brackets represent the word composition order in each slot.

\hspace{-1cm}
\tiny
\begin{tabular*}{\textwidth}{l l l l l l} 
\begin{tabular}{|p{2.1cm}|}
\hline
\textit{t=0} \\
\textit{input:} \\ \hline
\textless S\textgreater \\
A \\
little \\
child \\
sits \\
quietly \\
on \\
a \\
hand \\
built \\
rock \\
wall \\
in \\
autumn \\
\hline
\end{tabular}
&
\begin{tabular}{|p{2.1cm}|}
\hline
\textit{t=1} \\
\textit{input: \textless S\textgreater} \\ \hline
\textless S\textgreater \\
\cellcolor{maroon!20} (\textless S\textgreater A) \\
little \\
child \\
sits \\
quietly \\
on \\
a \\
hand \\
built \\
rock \\
wall \\
in \\
autumn \\
\hline
\end{tabular}
&
\begin{tabular}{|p{2.1cm}|}
\hline
\textit{t=2} \\
\textit{input: A} \\ \hline
\cellcolor{maroon!20}(A \textless S\textgreater) \\
(\textless S\textgreater A) \\
little \\
child \\
sits \\
quietly \\
on \\
a \\
hand \\
built \\
rock \\
wall \\
in \\
autumn \\
\hline
\end{tabular}
&
\begin{tabular}{|p{2.1cm}|}
\hline
\textit{t=3} \\
\textit{input: little} \\ \hline
(A \textless S\textgreater) \\
(\textless S\textgreater A) \\
little \\
child \\
sits \\
\cellcolor{maroon!20} (little quietly) \\
on \\
a \\
hand \\
built \\
rock \\
wall \\
in \\
autumn \\
\hline
\end{tabular}
&
\begin{tabular}{|p{2.1cm}|}
\hline
\textit{t=4} \\
\textit{input: child} \\ \hline
(A \textless S\textgreater) \\
(\textless S\textgreater A) \\
little \\
child \\
sits \\
\cellcolor{maroon!20} (child (little quietly))\\
on \\
a \\
hand \\
built \\
rock \\
wall \\
in \\
autumn \\
\hline
\end{tabular}
&
\begin{tabular}{|p{2.1cm}|}
\hline
\textit{t=5} \\
\textit{input: sits} \\ \hline
(A \textless S\textgreater) \\
(\textless S\textgreater A) \\
little \\
child \\
sits \\
(child (little quietly))\\
on \\
a \\
hand \\
built \\
rock \\
wall \\
in \\
\cellcolor{maroon!20} (sits autumn) \\
\hline
\end{tabular}

\\
\begin{tabular}{|p{2.1cm}|}
\hline
\end{tabular}\\
\begin{tabular}{|p{2.1cm}|}
\hline
\textit{t=6} \\
\textit{input: quietly} \\ \hline
(A \textless S\textgreater) \\
(\textless S\textgreater A) \\
little \\
child \\
\cellcolor{maroon!20}(quietly sits) \\
(child (little quietly))\\
on \\
a \\
hand \\
built \\
rock \\
wall \\
in \\
(sits autumn) \\
\hline
\end{tabular}
&
\begin{tabular}{|p{2.1cm}|}
\hline
\textit{t=7} \\
\textit{input: on} \\ \hline
(A \textless S\textgreater) \\
(\textless S\textgreater A) \\
little \\
child \\
(quietly sits) \\
(child (little quietly))\\
on \\
a \\
hand \\
built \\
rock \\
wall \\
in \\
\cellcolor{maroon!20}(on (sits autumn))\\
\hline
\end{tabular}
&
\begin{tabular}{|p{2.1cm}|}
\hline
\textit{t=8} \\
\textit{input: a} \\ \hline
(A \textless S\textgreater) \\
(\textless S\textgreater A) \\
little \\
child \\
(quietly sits) \\
(child (little quietly))\\
on \\
a \\
hand \\
\cellcolor{maroon!20}(a built) \\
rock \\
wall \\
in \\
(on (sits autumn))\\
\hline
\end{tabular}
&
\begin{tabular}{|p{2.1cm}|}
\hline
\textit{t=9} \\
\textit{input: hand} \\ \hline
(A \textless S\textgreater) \\
(\textless S\textgreater A) \\
little \\
child \\
(quietly sits) \\
(child (little quietly))\\
on \\
a \\
hand \\
(a built) \\
rock \\
\cellcolor{maroon!20}(hand wall) \\
in \\
(on (sits autumn))\\
\hline
\end{tabular}
&
\begin{tabular}{|p{2.1cm}|}
\hline
\textit{t=10} \\
\textit{input: built} \\ \hline
(A \textless S\textgreater) \\
(\textless S\textgreater A) \\
little \\
child \\
(quietly sits) \\
(child (little quietly))\\
on \\
a \\
hand \\
(a built) \\
rock \\
\cellcolor{maroon!20}(built (hand wall))\\
in \\
(on (sits autumn))\\
\hline
\end{tabular}
&
\begin{tabular}{|p{2.1cm}|}
\hline
\textit{t=11} \\
\textit{input: rock} \\ \hline
(A \textless S\textgreater) \\
(\textless S\textgreater A) \\
little \\
child \\
(quietly sits) \\
(child (little quietly))\\
on \\
a \\
hand \\
(a built) \\
rock \\
\cellcolor{maroon!20}(rock (built (hand wall)))\\
in \\
(on (sits autumn))\\
\hline
\end{tabular}
\\
\begin{tabular}{|p{2.1cm}|}
\hline
\end{tabular}\\
\begin{tabular}{|p{2.1cm}|}
\hline
\textit{t=12} \\
\textit{input: wall} \\ \hline
(A \textless S\textgreater) \\
(\textless S\textgreater A) \\
little \\
child \\
\cellcolor{maroon!20}(wall (quietly sits))\\
(child (little quietly))\\
on \\
a \\
hand \\
(a built) \\
rock \\
(rock (built (hand wall)))\\
in \\
(on (sits autumn))\\
\hline
\end{tabular}
&
\begin{tabular}{|p{2.1cm}|}
\hline
\textit{t=13} \\
\textit{input: in} \\ \hline
(A \textless S\textgreater) \\
(\textless S\textgreater A) \\
little \\
child \\
(wall (quietly sits))\\
(child (little quietly))\\
on \\
a \\
hand \\
(a built) \\
\cellcolor{maroon!20}(in rock) \\
(rock (built (hand wall)))\\
in \\
(on (sits autumn))\\
\hline
\end{tabular}
&
\begin{tabular}{|p{2.1cm}|}
\hline
\textit{t=14} \\
\textit{input: autumn} \\ \hline
(A \textless S\textgreater) \\
(\textless S\textgreater A) \\
little \\
child \\
(wall (quietly sits))\\
(child (little quietly))\\
on \\
a \\
hand \\
\cellcolor{maroon!20}(autumn (a built))\\
(in rock) \\
(rock (built (hand wall)))\\
in \\
(on (sits autumn))\\
\hline
\end{tabular}
\end{tabular*}

\end{document}